\documentclass[11pt]{article}

\usepackage[final]{acl}
\usepackage{algorithm}
\usepackage{algorithmic}
\usepackage{tikz}
\usepackage{forest}
\usepackage{tcolorbox}
\tcbuselibrary{breakable,skins}
\usepackage[table]{xcolor}

\newtcolorbox{datasetbox}[1][]{
    enhanced,
    breakable,
    colback=white,
    colframe=black!50,
    arc=0pt,
    outer arc=0pt,
    fonttitle=\bfseries,
    title=#1,
    boxrule=1pt,
    top=8pt,
    bottom=8pt,
    left=8pt,
    right=8pt
}

\usepackage{amsmath,amssymb,amsfonts}

\usepackage{graphicx}
\usepackage{subfigure}
\usepackage{booktabs}

\usepackage{times}
\usepackage{latexsym}

\usepackage[T1]{fontenc}

\usepackage[utf8]{inputenc}

\usepackage{microtype}

\usepackage{inconsolata}

\usepackage{graphicx}

\title{SMRC: Aligning Large Language Models with Student Reasoning for Mathematical Error Correction}

\author{
 \textbf{Biaojie Zeng}\textsuperscript{1,*},
 \textbf{Min Zhang}\textsuperscript{1,$\ddagger$,$\dagger$},
 \textbf{Juan Zhou}\textsuperscript{1,*},
 \textbf{Fengrui Liu}\textsuperscript{1},
 \textbf{Ruiyang Huang}\textsuperscript{2},
 \textbf{Yu Song}\textsuperscript{1},
 \textbf{Xin Lin}\textsuperscript{1,$\dagger$},
\\
 \textsuperscript{1}East China Normal University,
 \textsuperscript{2}Southeast University, 
\\
 \small \textsuperscript{*}Equal contribution \quad  \textsuperscript{$\ddagger$}Project leader 
 \quad 
 \href{52285901054@stu.ecnu.edu.cn}{52285901054@stu.ecnu.edu.cn} \quad \textsuperscript{$\dagger$}Corresponding authors: \href{mailto:mzhang@cs.ecnu.edu.cn}{mzhang@cs.ecnu.edu.cn}, \href{xlin@cs.ecnu.edu.cn}{xlin@cs.ecnu.edu.cn}
}

\begin{document}
\maketitle



\begin{abstract}
Large language models (LLMs) often make reasoning errors when solving mathematical problems, and how to automatically detect and correct these errors has become an important research direction. However, existing approaches \textit{mainly focus on self-correction within the model}, which falls short of the ``teacher-style” correction required in educational settings, \textit{i.e.}, systematically guiding and revising a student’s problem-solving process. To address this gap, we propose \texttt{SMRC} (\textit{\underline{S}tudent \underline{M}athematical \underline{R}easoning \underline{C}orrection}), a novel method that aligns LLMs with student reasoning. Specifically, \texttt{SMRC} formulates student reasoning as a multi-step sequential decision problem and introduces Monte Carlo Tree Search (MCTS) to explore optimal correction paths. To reduce the cost of the annotating process-level rewards, we leverage breadth-first search (BFS) guided by LLMs and final-answer evaluation to generate reward signals, which are then distributed across intermediate reasoning steps via a back-propagation mechanism, enabling fine-grained process supervision. Additionally, we construct a benchmark for high school mathematics, MSEB (Multi-Solution Error Benchmark), consisting of 158 instances that include problem statements, student solutions, and correct reasoning steps. We further propose a dual evaluation protocol centered on \textbf{solution accuracy} and \textbf{correct-step retention}, offering a comprehensive measure of educational applicability. Experiments demonstrate that \texttt{SMRC} significantly outperforms existing methods on two public datasets (ProcessBench and MR-GSM8K) and our MSEB in terms of effectiveness and overall performance. The code are
available at \url{https://github.com/ECNU-RAIL/SMRC-EMNLP2026}.
\end{abstract}

\begin{figure*}[!t]
  \centering  
  \includegraphics[width=0.96\textwidth]{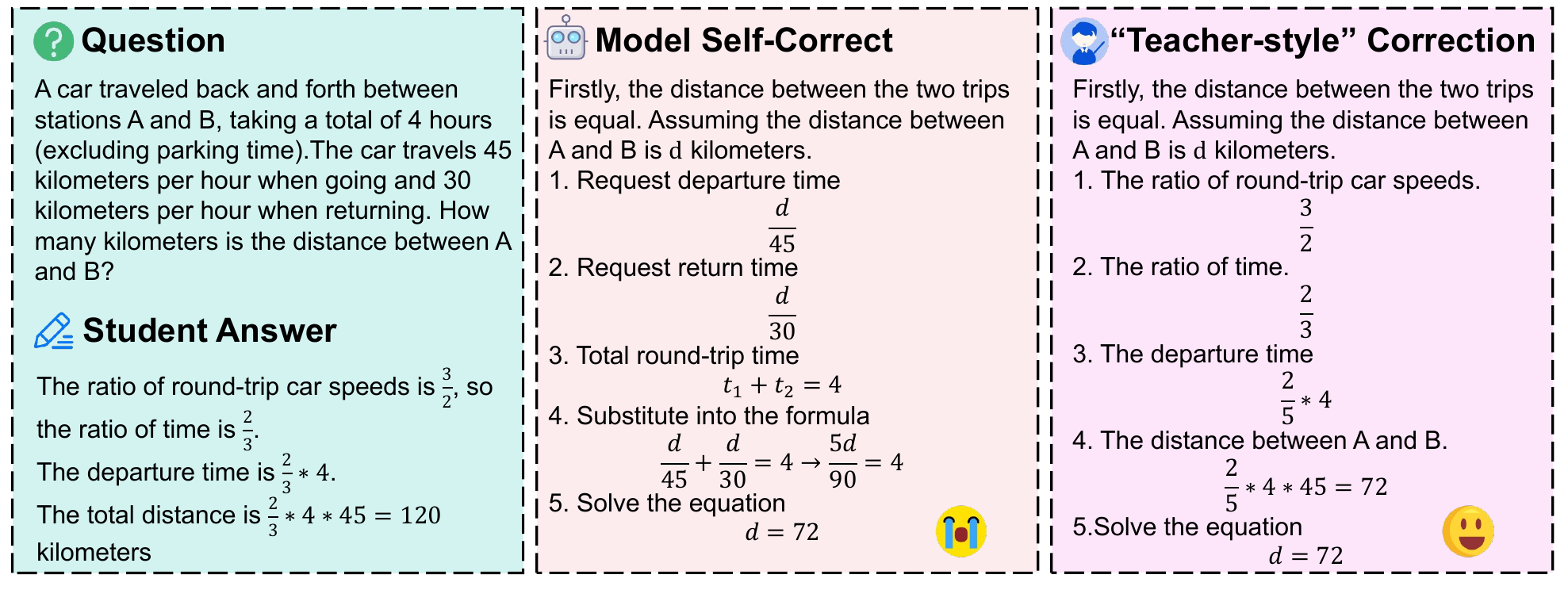}
  \caption{Example of two correction modes: given an input question and student answer (Left), model self-correction detects reasoning errors in LLMs (Middle), while ``teacher-style" correction identifies errors in the student's reasoning process (Right).}
  \label{fig:teaser}
\end{figure*}

\section{Introduction}


In recent years, large language models (LLMs) have achieved remarkable progress in complex mathematical reasoning tasks, thanks to their powerful logical and computational abilities~\citep{li2025limrrlscaling}. These advances have not only improved the performance of LLMs in scientific and engineering domains but have also accelerated their integration into educational applications. However, existing studies reveal that even state-of-the-art reasoning models remain prone to logical inconsistencies and computational errors~\citep{Boye2025LargeLM}. To address these challenges, researchers have proposed self-correction approaches for LLMs. By incorporating iterative refinement, multi-candidate verification, and consistency checks, these methods move beyond traditional single-pass generation and establish a unified generate–verify–correct framework, enabling models to autonomously detect and rectify internal reasoning errors, thereby improving the reliability of mathematical inference.

Meanwhile, the applications of LLMs in education have become increasingly diverse~\citep{liu2024socraticlm, MuduoLLM2025}. Beyond correcting their own reasoning errors, LLMs are now being leveraged as “teacher-like” agents to identify and correct students’ reasoning mistakes in problem-solving. However, these two types of correction differ fundamentally in both goals and mechanisms, as illustrated in Figure~\ref{fig:teaser}. Model self-correction focuses on enhancing logical consistency within the model’s reasoning, whereas student-oriented correction emphasizes pedagogical coherence and cognitive support~\citep{li2024improving,  Narciss2004HowTD}. In educational contexts, effective correction involves not only providing the correct final answer but, more importantly, maintaining continuity with the student’s original reasoning path to foster deeper cognitive development. Recent benchmark datasets, such as ProcessBench~\citep{processbench} and MR-GSM8K~\citep{zeng2023challenge}, represent initial steps in this direction. Nonetheless, these datasets rely on synthetically generated student responses, which fail to capture the nuanced cognitive patterns and systematic misconceptions of real learners. This limitation becomes particularly evident in practical teaching scenarios—models trained on simulated data often struggle to generalize to authentic student reasoning. Therefore, there is an urgent need for a correction framework grounded in real student data that simultaneously ensures mathematical rigor and pedagogical alignment, paving the way toward more effective human–AI collaborative teaching.

To overcome these limitations in existing research, we propose \texttt{SMRC} (\underline{S}tudent \underline{M}athematical \underline{R}easoning \underline{C}orrection), a novel framework designed to align LLMs with students’ reasoning processes. Specifically, \texttt{SMRC} formulates student problem-solving as a multi-step sequential decision-making task and employs Monte Carlo Tree Search (MCTS) to explore optimal correction paths. To reduce the cost of the annotating process-level rewards, we leverage LLM-guided breadth-first search (BFS) in conjunction with final-answer evaluation to generate reward signals, which are then back-propagated across intermediate reasoning steps to enable fine-grained process supervision. Additionally, we construct a high school mathematics benchmark dataset, termed \textbf{MSEB (Multi-Solution Error Benchmark), comprising 158 authentic student solution records.} Unlike existing benchmarks that rely on model-simulated errors, MSEB preserves the complete reasoning trajectories of students, capturing their genuine cognitive patterns and common error types. Each problem in MSEB includes multiple valid solution strategies, enabling comprehensive evaluation of correction methods in terms of both maintaining mathematical correctness and preserving the student’s original reasoning logic. Our contributions are summarized as follows:

\begin{itemize}
    \item We introduce the novel research problem of Student Mathematical Reasoning Correction (\texttt{SMRC}), emphasizing that in educational contexts, correction should not only ensure mathematical accuracy but also preserve students’ original reasoning paths to support cognitive development.
    \item \texttt{SMRC} employs Monte Carlo Tree Search to explore optimal correction paths and combines LLM-guided breadth-first search with back-propagated process-level rewards to achieve fine-grained supervision, effectively balancing correction accuracy with fidelity to students’ thoughts.
    \item We construct the MSEB dataset, comprising 158 authentic high-school student solution records. The dataset preserves complete reasoning paths and includes multiple valid solution methods, enabling comprehensive evaluation of correction approaches in terms of both mathematical correctness and preservation of student reasoning. Extensive experiments validate the effectiveness of our method.
\end{itemize}

\section{Related Work}

\textbf{Self-correction for LLMs.} Self-correction has emerged as a critical approach for improving reasoning accuracy in large language models. Recent datasets such as ProcessBench~\citep{processbench} and MR-GSM8K~\citep{zeng2023challenge} use open-source models to generate erroneous reasoning patterns that simulate student problem-solving. Representative methods include Self-Check~\citep{miao2024selfcheck}, which employs internal consistency verification, and Self-Refine~\citep{madaan2023selfrefine}, which uses iterative refinement strategies. However, existing approaches optimize primarily for answer correctness while neglecting a key educational requirement: preserving students' original reasoning strategies. In real-world tutoring, effective correction must build upon students' initial approaches rather than replacing them entirely. This gap motivates our work—developing student-centered correction that maintains fidelity to original reasoning while achieving high accuracy.

\textbf{Test-time scaling (TTS).} 
As pretraining-phase scaling has plateaued, test-time scaling has emerged as a key research focus. By reallocating computational resources from training to inference, TTS improves performance without additional training. Representative approaches include: (1) Parallel Scaling: generating $k$ candidates and selecting the best via reward models~\citep{wang2023selfconsistency, sun2024easytohard}; (2) Tree Search: using reward-guided exploration to find high-quality reasoning paths~\citep{zhang2023planning, Wang2024LiteSearchET}; (3) Long Chain-of-Thought: producing extended reasoning sequences with iterative verification~\citep{Xia2025TokenSkipCC, Yang2025TowardsTS}; (4) Multi-round Refinement: iteratively improving outputs through self-correction. Our work adopts Monte Carlo Tree Search (MCTS) for mathematical error correction, using external reward models to compute $Q$-values that guide path selection and prevent inherent model biases. This design balances thought preservation with correction accuracy~\citep{welleck2023generating, Xi2024EnhancingLR}.

\textbf{Reward models (RMs).} 
Reward models guide model outputs toward human preferences by converting them into optimizable scalar signals~\citep{Gao2022ScalingLF}. They are categorized into two types~\citep{Wang2025ASO}: Outcome-supervised Reward Models (ORMs) evaluate final answers, while Process-supervised Reward Models (PRMs) assess intermediate reasoning steps~\citep{Zhong2025ACS}. These models enable quality optimization during training or inference~\citep{Zeng2023OnDP}.
Unlike traditional reward models, our approach introduces a specialized reward model for student mathematical reasoning correction that identifies optimal nodes where students achieve maximum reasoning progress. To address the scarcity of process-supervised labels, we develop an algorithm that infers process-level supervision from outcome labels through reasoning tree construction and label propagation. This enables precise guidance within the \texttt{SMRC} framework, optimizing for both mathematical correctness and consistency with students' original reasoning approaches.

\section{Preliminaries}

\label{sec:preliminaries}

In this section, we will briefly and separately introduce the problem definition of self-correction for LLMs and the Monte Carlo Tree Search method.

\subsection{Self-Correction for LLMs}

Self-correction for LLMs aims to enable models to autonomously identify and correct errors in their mathematical reasoning processes. In recent years, researchers have proposed various self-correction methods, such as iterative self-refinement strategies and internal consistency verification mechanisms~\citep{madaan2023selfrefine, miao2024selfcheck}, driving significant research interest in this area. We formalize this task as:

Let $Q$ denote a mathematical problem and $A = \{a_1, a_2, \ldots, a_n\}$ represent a reasoning sequence that may contain logical errors or miscalculations. The objective of LLM self-correction specifically aims to generate a corrected sequence $A^*$ that satisfies:
\begin{equation}
    A^* = \arg\max_{A'} \underbrace{P(A' \vdash Q_{\text{answer}} | Q, A)}_{\text{Correctness Probability}}
    \label{eq:scmr_objective_simple}
\end{equation}

\begin{figure*}
    \centering
    \includegraphics[width=0.965\linewidth]{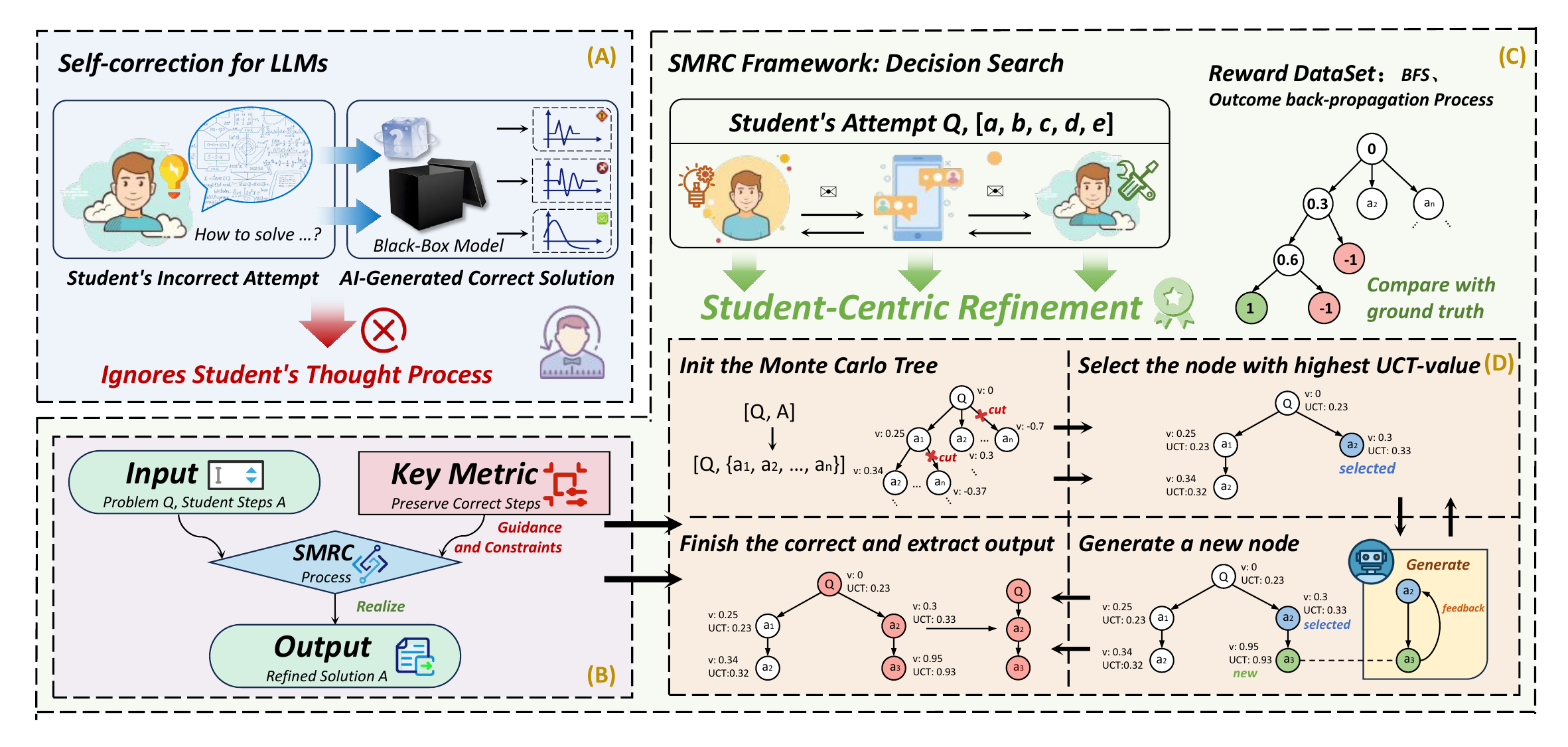}
    \caption{(A) Self-correction methods for LLMs often fail to address students' reasoning processes, relying instead on direct model-generated answers
(B) \texttt{SMRC} inputs student questions and initial attempts, using Monte Carlo Tree Search to generate educational corrections;
(C) The framework combines outcome and process rewards to train the reward model
(D) This method starts from student attempts and provides final corrections through four reasoning steps.}
    \label{fig:q-value}
\end{figure*}

In this formulation, $P(\cdot)$ denotes the probability function, $\vdash$ denotes logical entailment, and $Q_{\text{answer}}$ represents the correct answer to problem $Q$. Given reasoning sequence $A$, the objective is to find $A^*$ most likely to correctly derive the solution.

\subsection{Monte Carlo Tree Search (MCTS)}

Monte Carlo Tree Search (MCTS)~\citep{swiechowski2023monte} provides a principled framework for decision-making in complex search spaces through balanced exploration and exploitation. The algorithm iteratively constructs a search tree via four key phases: \textit{selection}, \textit{expansion}, \textit{simulation}, and \textit{backpropagation}.

During selection, the algorithm iteratively selects the child maximizing the UCT criterion:
\begin{equation}
s^* = argmax_{s \in node} \text{UCT}(s)
\label{eq:selection}
\end{equation}
where the UCT value is defined as:
\begin{equation}
\text{UCT}(s) = \frac{Q(s)}{N(s)} + c \sqrt{\frac{\ln N(s_{\text{parent}})}{N(s)}}
\label{eq:uct_formula}
\end{equation}
Here, $Q(s)$ denotes the cumulative reward of node $s$, $N(s)$ its visit count, and $c$ the exploration coefficient. The term $\frac{Q(s)}{N(s)}$ represents exploitation, favoring nodes with higher average rewards, while $c\sqrt{\frac{\ln N(s_{\text{parent}})}{N(s)}}$ represents exploration, encouraging the search of less frequently visited nodes.

In the expansion phase, new child nodes are added, followed by rollout-based simulation to obtain a reward and backpropagation to update node statistics. This process gradually concentrates search on the most promising regions.

\section{Methodology}
While existing self-correction frameworks in Section \ref{sec:preliminaries} emphasize answer correctness, they overlook the preservation of students’ reasoning, which is crucial in educational settings. We propose Student Mathematical Reasoning Correction (\texttt{SMRC}), which formulates error correction as a sequential decision-making problem and applies Monte Carlo Tree Search (MCTS) with a specialized reward function to improve answer accuracy while maintaining alignment with students’ reasoning trajectories.

\subsection{Problem Formulation: Student-Aware Mathematical Reasoning Correction}
\label{subsec:problem_formulation}

Self-correction tasks for large language models have primarily focused on mathematical reasoning correctness. However, in educational contexts, correctness alone is insufficient: abrupt changes in reasoning may confuse students and hinder learning. We therefore define a new task, \emph{\textbf{student-aware mathematical reasoning correction}}, which requires corrected solutions to be both accurate and consistent with students' reasoning processes.

Although this task may appear similar to conventional self-correction, it differs fundamentally in its objective. The corrected reasoning sequence $A^*$ must not only yield the correct answer to a problem $Q$, but also preserve valid steps from the student's original reasoning sequence $A$. This requires explicitly balancing solution correctness with fidelity to the student's cognitive trajectory.

Formally, the objective is defined as:
\begin{equation}
\begin{split}
A^* 
&= \arg\max_{A'} P\!\left(A' \vdash Q_{\text{answer}} \mid Q, A\right) \\
&\quad + \lambda \sum_{i=1}^{n} \mathbb{I}\!\left(\text{preserve}(s_i, A')\right),
\end{split}
\label{eq:SAMRC_objective}
\end{equation}

where $\mathbb{I}(\cdot)$ is an indicator function that equals 1 if a correct reasoning step 
$s_i \in A_{\text{correct}} \subseteq A$ is preserved in $A'$, and 0 otherwise.

\subsection{\texttt{SMRC} Framework Overview}
\label{subsec:framework_overview}

To correct mathematical errors while preserving students' original reasoning patterns, we propose the \texttt{SMRC} framework (Fig.~\ref{fig:q-value}). The framework consists of two core components: a reward model and Monte Carlo Tree Search (MCTS). In the following, we describe each component in detail.

\begin{figure*}[ht]
    \centering
    \includegraphics[width=0.965\linewidth]{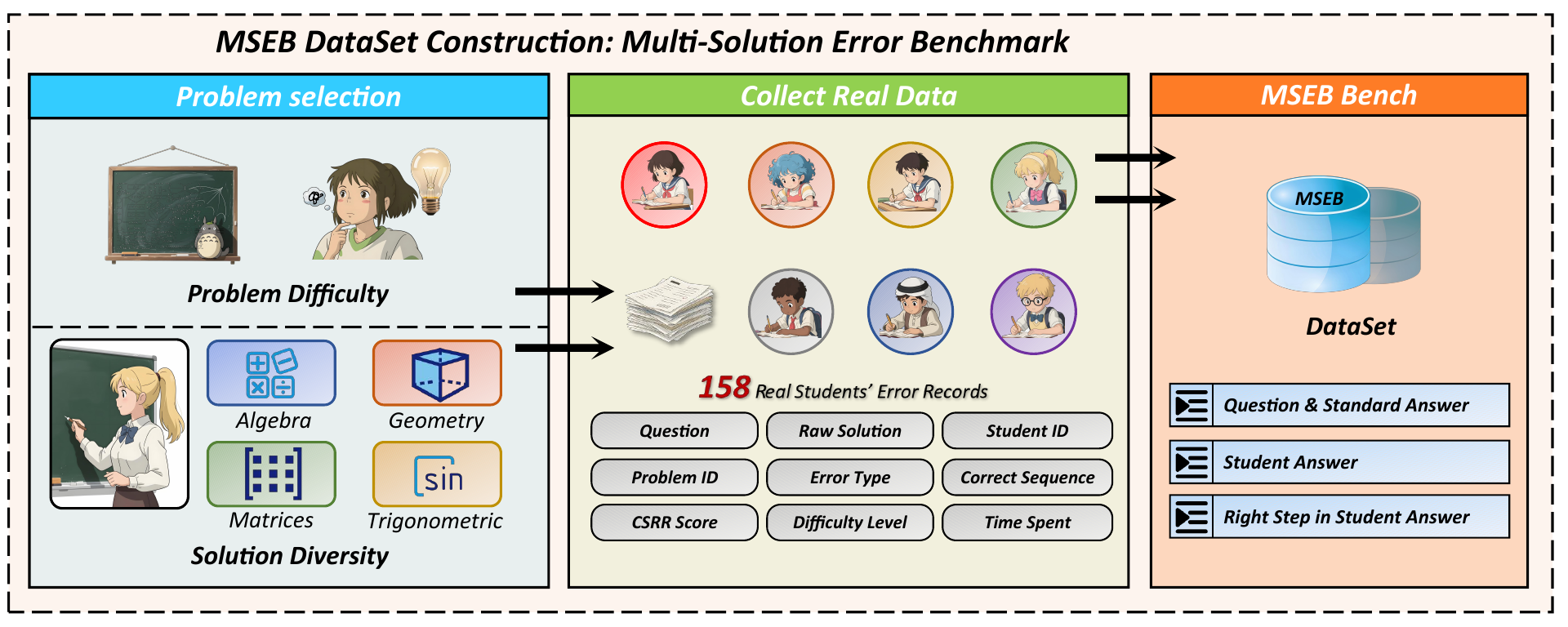}
    \caption{MSEB Dataset Construction Pipeline.}
    \label{fig:datasets}
\end{figure*}

\subsubsection{Reward Model in \texttt{SMRC}}
\label{subsubsec:reward}

In \texttt{SMRC}, the reward model outputs are treated as Q-values to guide node selection in MCTS. Since \textbf{fine-grained process supervision is extremely scarce}, we propose a \textbf{difference-based backtracking algorithm} that transforms sparse outcome-level labels into dense process-level supervision signals. The algorithm consists of two steps:

\textbf{Reasoning Tree Construction:} We construct reasoning trees using LLM-based BFS. The root node is initialized with value $0$, while leaf nodes are assigned $\pm 1$ according to answer correctness.

\textbf{Differential Reward Allocation:} 
For each leaf node, we backtrack along its path to the nearest ancestor with an assigned reward, compute the reward difference, and uniformly distribute it across the nodes in the corresponding path segment. The final reward of each node is obtained by cumulative aggregation, reflecting the overall quality of the reasoning path from the root to that node.

The reward model trained with these process-level supervision signals accurately evaluates the contribution of each reasoning step. \textbf{Its outputs are directly used as Q-value estimates in MCTS}, providing core guidance for node selection.
A construction example is provided in the appendix \ref{app:reward}.

\subsubsection{Monte Carlo Tree Search in \texttt{SMRC}}
\label{subsubsec:MCTS}
To systematically explore and optimize reasoning path in student problem-solving processes, \texttt{SMRC} employs the Monte Carlo Tree Search (MCTS) to locate errors and generate correction solutions. The specific process is as follows:

\textbf{Tree Initialization:} 
The student’s solution is decomposed into 
$n$ atomic reasoning steps. Starting from the root node (empty path), the tree is built layer by layer: the first layer contains single steps, and later layers extend paths by adding unused steps in their original order. Enumerating all ordered subsets yields a complete tree with $2^n$
 nodes. An example is shown in Appendix \ref{appedix:tree_case}.

\begin{algorithm}[t]
\caption{Student Mathematical Reasoning Correction (\texttt{SMRC})}
\label{alg:SMRC}
\begin{algorithmic}[1]
\REQUIRE Math problem $Q$, Student's attempt $A$, Reward model $R$, Maximum attempts $T$, Reward threshold $\theta$
\ENSURE Corrected solution $A^*$

\STATE $\mathcal{T} \gets \text{InitializeTree}(Q, A, R)$

\FOR{$t = 1$ \TO $T$}
    \STATE $s_{\text{select}} \gets \text{SelectNode}(\mathcal{T})$
    \STATE $s_{\text{new}} \gets \text{GenerateValidatedStep}(s_{\text{select}})$
    \IF{$\text{GenerationFailed}(s_{\text{new}})$}
        \STATE \textbf{continue}
    \ENDIF

    \STATE $q_{\text{new}} \gets R(s_{\text{new}})$
    \IF{$q_{\text{new}} \leq s_{\text{select}}.Q$}
        \STATE \textbf{continue}
    \ENDIF

    \STATE $\text{AddNode}(s_{\text{select}}, s_{\text{new}}, q_{\text{new}}) $
    \STATE $\text{UpdateAncestor}(s_{\text{new}}, q_{\text{new}})$

    \IF{$q_{\text{new}} \geq \theta$}
        \RETURN $\text{GetPathToRoot}(s_{\text{new}})$
    \ENDIF
\ENDFOR

\STATE $s_{\text{best}} \gets \text{FindHighestScoringNode}(\mathcal{T})$
\RETURN $\text{GetPathToRoot}(s_{\text{best}})$
\end{algorithmic}
\end{algorithm}

\textbf{Node Selection:} The reward model computes Q-values for each node to evaluate the quality of its reasoning path. Based on Q-values, the Upper Confidence Bound for Trees (UCT) criterion is used to balance exploration and exploitation, selecting the most promising student reasoning node as the basis for subsequent reasoning.

\textbf{Reasoning Path Generation and Validation:} 
Given the reasoning path of the selected node, the large language model generates subsequent steps to extend the student's solution. A reward model evaluates the newly generated nodes, and those that fail to yield positive improvement over the selected node are pruned, retaining only high-value branches to improve search efficiency.

\textbf{Backpropagation and Output:} 
The reward of a newly generated reasoning node is backpropagated to update the Q-values of its ancestor nodes. The algorithm terminates when a complete reasoning path exceeds a predefined reward threshold or the maximum number of iterations is reached, and returns the highest-reward path as the final solution.

\texttt{SMRC} turns error correction into a constrained reasoning task, selecting the best node via a reward model to preserve correct steps and improve correction accuracy. The complete algorithmic workflow is described in Algorithm \ref{alg:SMRC}.

\begin{table*}[t]
  \centering
  \resizebox{0.965\textwidth}{!}{
  \begin{tabular}{l|cc>{\columncolor{blue!20}}ccc>{\columncolor{blue!20}}ccc>{\columncolor{blue!20}}c}
    \toprule
     &\multicolumn{3}{c}{ProcessBench} & \multicolumn{3}{c}{MR-GSM8K} & \multicolumn{3}{c}{MSEB} \\
    \midrule
    Method & ACC & CSRR & HM & ACC & CSRR & HM & ACC & CSRR & HM \\
    
    \midrule

    \rowcolor{gray!20} \multicolumn{10}{c}{Qwen2.5-7B-Instruct} \\ 
      DEC & 50.7 $\pm$ 0.4 & 67.7 $\pm$ 0.2 & 58.0 $\pm$ 0.2 & 60.5 $\pm$ 0.9 & 79.6 $\pm$ 0.2 & 68.7 $\pm$ 0.7 & 18.7 $\pm$ 1.1 & 56.4 $\pm$ 1.8 & 28.1 $\pm$ 1.0 \\
      TPEC & 53.6 $\pm$ 0.4 & 92.9 $\pm$ 0.1 & 68.0 $\pm$ 0.3 & 63.1 $\pm$ 0.5 & 94.7 $\pm$ 0.2 & 75.7 $\pm$ 0.4 & 17.9 $\pm$ 2.6 & 84.7 $\pm$ 1.1 & 29.4 $\pm$ 3.3 \\

    \rowcolor{gray!20} \multicolumn{10}{c}{Llama-3.1-8B-Instruct} \\ 
      DEC & 30.9 $\pm$ 3.7 & 66.0 $\pm$ 1.5 & 42.7 $\pm$ 3.2 & 37.4 $\pm$ 0.7 & 81.1 $\pm$ 0.5 & 51.1 $\pm$ 0.6 & 6.75 $\pm$ 1.1 & 38.3 $\pm$ 2.2 & 11.4 $\pm$ 1.4\\
      TPEC & 29.7 $\pm$ 2.8 & 80.8 $\pm$ 0.7 & 43.4 $\pm$ 2.1& 37.4 $\pm$ 0.5 & 92.0 $\pm$ 0.5 & 53.2 $\pm$ 0.5 & 8.03 $\pm$ 2.1 & 65.2 $\pm$ 2.8 & 14.2 $\pm$ 3.3\\

    \rowcolor{gray!20} \multicolumn{10}{c}{Qwen2.5-14B-Instruct} \\ 
      DEC & 56.8 $\pm$ 0.4 & 93.6 $\pm$ 0.2 & 70.7 $\pm$ 0.3 & 76.1 $\pm$ 0.6 & 88.8 $\pm$ 0.2 & 82.0 $\pm$ 0.4 & 21.4 $\pm$ 1.6 & 81.7 $\pm$ 1.1 & 34.0 $\pm$ 2.0 \\
     TPEC & 57.9 $\pm$ 0.2 & \textbf{96.3 $\pm$ 0.1} & 72.3 $\pm$ 0.1 & 80.7 $\pm$ 0.2 & 93.5 $\pm$ 0.2 & 86.7 $\pm$ 0.2 & 25.0 $\pm$ 0.6 & \underline{88.2 $\pm$ 0.9} & 39.1 $\pm$ 0.8 \\

    \rowcolor{gray!20} \multicolumn{10}{c}{MuduoLLM} \\ 
     DEC & 52.0 $\pm$ 0.1 & 30.3 $\pm$ 0.2 & 38.3 $\pm$ 0.1 & 21.4 $\pm$ 0.5 & 25.4 $\pm$ 0.3 & 23.2 $\pm$ 0.4 & 17.1 $\pm$ 0.9 & 28.3 $\pm$ 1.2 & 21.3 $\pm$ 1.0\\
     TPEC & 56.8 $\pm$ 0.7 & 84.0 $\pm$ 0.2 & 67.8 $\pm$ 0.6 & 71.9 $\pm$ 0.5 & 86.4 $\pm$ 0.3 & 78.5 $\pm$ 0.4 & 19.8 $\pm$ 2.4 & 57.0 $\pm$ 2.5 & 29.3 $\pm$ 2.7\\

    \rowcolor{gray!20} \multicolumn{10}{c}{Llama-3.1-70B-Instruct} \\ 
      DEC & 48.3 $\pm$ 0.1 & 29.2 $\pm$ 0.3 & 36.4 $\pm$ 0.2 & 52.6 $\pm$ 1.2 & 46.7 $\pm$ 0.6 & 49.5 $\pm$ 0.8 & 10.4 $\pm$ 0.6 & 27.6 $\pm$ 2.8 & 15.0 $\pm$ 1.0\\
      TPEC & 41.0 $\pm$ 0.4 & 78.5$\pm$ 0.5 & 53.8 $\pm$ 0.3 & 72.5 $\pm$ 0.8 & 91.3 $\pm$ 0.3 & 80.9 $\pm$ 0.4 & 11.3 $\pm$ 0.9 & 82.3 $\pm$ 0.8 & 20.0 $\pm$ 1.4\\
    
     \rowcolor{gray!20} \multicolumn{10}{c}{Qwen2.5-72B-Instruct} \\ 
      DEC & \underline{64.4 $\pm$ 0.5} & 82.1 $\pm$ 0.2 & 72.1 $\pm$ 0.3 & 78.7 $\pm$ 0.7 & 91.0 $\pm$ 0.3 & 84.4 $\pm$ 0.5 & 26.5 $\pm$ 0.5 & 69.4 $\pm$ 3.5 & 38.3 $\pm$ 0.9\\
      TPEC & 61.6 $\pm$ 0.9 & 92.5 $\pm$ 5.3 & 73.9 $\pm$ 2.4 & 82.2 $\pm$ 0.4 & \textbf{96.4 $\pm$ 0.1} & 88.7 $\pm$ 0.2 & 27.2 $\pm$ 1.6 & 82.1 $\pm$ 0.8 & 40.8 $\pm$ 1.7\\

    \midrule
    Self-Check &62.5 $\pm$ 0.3 & \underline{95.3 $\pm$ 0.6} & \textbf{75.5 $\pm$ 0.3} & 83.9 $\pm$ 0.3 & \underline{95.2 $\pm$ 0.1} & \underline{89.2 $\pm$ 0.2} & 27.2 $\pm$ 0.1 & \textbf{91.2 $\pm$ 1.6} & 41.9 $\pm$ 1.1\\
    Self-Refine & 60.3 $\pm$ 0.4 & 85.9 $\pm$ 0.9 & 70.8 $\pm$ 0.6 & \underline{86.9 $\pm$ 1.5} & 90.4 $\pm$ 0.6 & 88.6 $\pm$ 0.5 & \underline{35.2 $\pm$ 2.1} & 76.6 $\pm$ 4.2 & \underline{48.1 $\pm$ 1.8} \\
    \texttt{SMRC} (ours)  & \textbf{64.9 $\pm$ 0.4} & 89.3 $\pm$ 0.4 & \underline{75.2 $\pm$ 0.2} & \textbf{91.4 $\pm$ 0.5} & 94.5 $\pm$ 0.2 & \textbf{92.9 $\pm$ 0.2} & \textbf{40.1 $\pm$ 1.7} & 73.7 $\pm$ 2.1 & \textbf{51.8 $\pm$ 1.1}\\
    \bottomrule
  \end{tabular}}
  \caption{Results of \texttt{SMRC} and other frontier LLMs on three benchmarks. Methods without model specification use Qwen2.5-72B-Instruct. \textbf{Bold} denotes the best, and \underline{underlined} the second-best results.}
  \label{tab:results}
\end{table*}

\subsection{MSEB Dataset Construction}
To evaluate \texttt{SMRC}, we construct a student problem-solving dataset, \textbf{MSEB (Multi-Solution Error Benchmark)}. Unlike prior work relying on model-simulated student errors, MSEB is derived from high school students' problem-solving processes, providing a reliable benchmark for fine-grained student error identification and correction.

\textbf{Problem Selection:} 
We select problems from a high school mathematics question bank covering algebra, geometry, matrices, and trigonometry, guided by two criteria:
\textbf{(i) Problem Difficulty}, ensuring sufficient complexity to elicit diverse reasoning and common student errors; and
\textbf{(ii) Solution Diversity}, selecting problems with multiple valid solution paths to assess whether correction methods preserve students’ original correct reasoning.

\textbf{Data Collection and Annotation:}
We collect complete written solutions from high school students solving the selected problems. After systematic organization and annotation, we construct a dataset of \textbf{158 real student error records}. Each record contains the original problem, standard answer(s), the student’s full solution, and the correctly reasoned steps. 

Example instances are provided in Appendix~\ref{app:data_ins}. 
Additional dataset statistics and coverage analyses are reported in Appendix~\ref{sec:mseb_analysis}. 
All data are anonymized to remove personally identifiable or sensitive information.

\section{Experiments}

In this section and appendix, we conduct experiments to evaluate the effectiveness of our proposed \texttt{SMRC} method, aiming to answer the following questions:
(1) Can \texttt{SMRC} significantly improve the model's ability in student reasoning correction? (Section \ref{sec:mainresult})
(2) Can instruct models directly serve as reward models? (Section \ref{sec:base_fine})
(3) How effective is \texttt{SMRC} in different settings (or ablation study)? (Section \ref{sec:feedback} and Appendix \ref{sec:c})
(4) How universal is \texttt{SMRC}, i.e., can it be directly plugged into various base models to achieve consistent performance improvements? (Appendix \ref{sec:gener})
(5) What is the computational cost of \texttt{SMRC} and the efficiency trade-off introduced by MCTS? (Appendix \ref{sec:efficiency})

\subsection{Experimental Setup}
\label{sec:setting}
\textbf{Benchmarks.}
We evaluate \texttt{SMRC} framework on three datasets: ProcessBench~\citep{processbench}, MR-GSM8K~\citep{zeng2023challenge}, and MSEB dataset. ProcessBench is built upon GSM8K~\citep{cobbe2021gsm8k}, MATH~\citep{hendrycksmath2021}, OlympiadBench~\citep{he-etal-2024-olympiadbench}, and Omni-MATH~\citep{gao2025omnimath}, while MR-GSM8K is based on GSM8K. Both public datasets use LLM-generated student reasoning with human-annotated error steps.
For CSRR evaluation, we treat all steps before the first annotated error as correct, since ProcessBench and MR-GSM8K datasets only label the first erroneous step.

\textbf{Baselines.}
We compare our proposed \texttt{SMRC} framework with the following baseline method: \textbf{Direct Error Correction (DEC)}: Directly prompt LLMs to correct student' incorrect answer. \textbf{Thought-Preserving Error Correction (TPEC)}: Based DEC and instruct LLMs to preserve student' original reasoning method during correction. \textbf{Self-Refine}~\citep{madaan2023selfrefine}: Optimizing reasoning path through multiple rounds of self-improvement. \textbf{Self-Check}~\citep{miao2024selfcheck}: Generates multiple reasoning paths and selects the output with the highest self-check score.
For DEC and TPEC methods, we evaluate on multiple base models, including open-source models of different scales~\citep{qwen2.5, Dubey2024TheL3} and education-specific models~\citep{MuduoLLM2025}. Self-Refine and Self-Check methods are evaluated using the Qwen2.5-72B-Instruct model.

\textbf{Implementation details.} We utilize Qwen2.5-14B-Instruct fine-tuned on our reward dataset to estimate Q-values, while Qwen2.5-72B-Instruct serves as the primary generator for node expansion in the Monte Carlo tree. The remaining hyperparameter settings are detailed in Appendix \ref{app:implement}.

\textbf{Metrics.} To evaluate the efficacy of our approach, we adopt three key performance metrics: 

(1) \textbf{Reasoning Accuracy (ACC):} 
\begin{equation}
    ACC = \frac{1}{|\mathcal{D}|} \sum_{i=1}^{|\mathcal{D}|} \mathbb{I}[\text{IsValid}(A_i^*, Q_i, RA_i)]
\end{equation} 

(2) \textbf{Correct Step Retention Rate (CSRR):}
\begin{equation}
CSRR = \frac{1}{|\mathcal{D}|} \sum_{i=1}^{|\mathcal{D}|} \frac{|A_i^* \cap A_i^{\text{correct}}|}{|A_i^{\text{correct}}|}
\end{equation}
The two metrics measure error correction accuracy and correct step retention, respectively, where $|D|$ denotes the dataset size, $A_i^*$ is the corrected solution, $Q_i$ is the $i$-th question, $RA_i$ is the reference answer, $A_i^{correct}$ is the set of correct steps in the original solution, and $IsValid(\cdot)$ is the validation function. We adopt the harmonic mean \textbf{(HM)} as a unified metric to jointly evaluate, defined as:
$$
HM = \frac{2 \cdot CSRR \cdot ACC}{CSRR + ACC}
$$
Following prior work~\citep{Ho2025LLMasaJudgeRT, zhang-etal-2025-lessons}, we use Qwen2.5-72B-Instruct~\citep{qwen2.5} with structured prompts for evaluation.

\begin{figure*}
    \centering
    \includegraphics[width=0.96\linewidth]{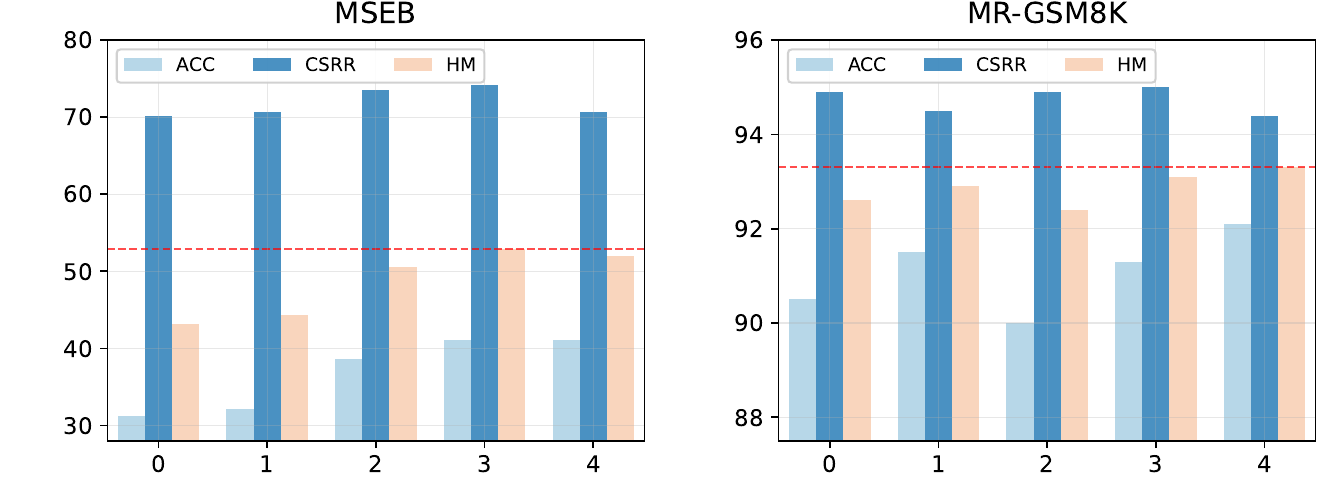}
    \caption{Impact of feedback iterations on performance in \texttt{SMRC}. The red line denotes the maximum HM value.}
    \label{fig:a1}
\end{figure*}

\begin{table*}[ht]
  \centering
  \resizebox{0.965\textwidth}{!}{
      \begin{tabular}{l|c|cc>{\columncolor{blue!10}}ccc>{\columncolor{blue!10}}ccc>{\columncolor{blue!10}}c>{\columncolor{blue!30}}c}
        \toprule
        & &\multicolumn{3}{c}{ProcessBench} & \multicolumn{3}{c}{MR-GSM8K} & \multicolumn{3}{c}{MSEB} \\
        \midrule
        Reward Model &Size &ACC & CSRR & HM & ACC & CSRR & HM & ACC & CSRR & HM & Avg. HM\\
        \midrule
        Qwen2.5-Instruct & 7B & 50.2 & 76.8& 60.7& 48.0 & 88.3 & 62.2 & 23.4 & 62.1 & 34.1 & 52.3\\
        
        Qwen2.5-Instruct & 14B & \underline{58.2} & \underline{84.4} & \underline{68.9} &  52.1 & 91.0 & 66.3   & \underline{26.5} & \textbf{75.1}& \underline{39.2} & \underline{58.1}\\
        
        Qwen2.5-Instruct &72B& 52.6 & 75.3 & 62.0 & 52.1 & 91.1 & 66.3 & 22.7 & 67.7 & 34.1 & 54.1 \\

        Llama-3.1-Instruct& 8B & 44.3 & 79.5 & 56.9 & \underline{61,9} & \underline{91.7} & \underline{73.9} & 16.4 & 61.1 & 25.9 & 52.2\\
        
        Llama-3.1-Instruct & 70B & 46.5 & 80.3 & 58.9 & 36.5 & 88.9&  51.7 & 20.2 & 72.7 & 31.6 & 47.4\\
        \midrule
        
        \texttt{SMRC} (ours) & 14B & \textbf{64.9 } & \textbf{89.3}  & \textbf{75.2} & \textbf{91.4} & \textbf{94.5} & \textbf{92.9} & \textbf{40.1 } & \underline{73.7}  & \textbf{51.8} & \textbf{73.3}\\
        \bottomrule
        \end{tabular}}

    \caption{Results of using open-source foundation models directly as reward models through prompt engineering. Among these, our proposed \texttt{SMRC} method, which is obtained by fine-tuning the Qwen2.5-14B model, demonstrates performance advantages. \textbf{Bold} indicates the best performance, and \underline{underline} indicates the second-best performance.}
  \label{tab:reward}
\end{table*}

\subsection{Main Results}
\label{sec:mainresult}

Table \ref{tab:results} presents a comprehensive performance comparison between our \texttt{SMRC} framework and baseline methods across three datasets. The experimental results reveal several key findings:

\textbf{\texttt{SMRC} shows versatility and robustness.}
It performs consistently across multiple error correction datasets. Although not the top method on ProcessBench, which emphasizes complex reasoning, \texttt{SMRC} achieves leading results on MR-GSM8K and MSEB. These results suggest strong adaptability to diverse correction settings and good generalization.

\texttt{SMRC} addresses the limitations of generation-based filtering through structured search. Guided by an external reward model, its MCTS framework explores correction paths more thoroughly and avoids local optima compared with methods such as Self-Check and Self-Refine. The advantage is especially evident on datasets like \textbf{MSEB}, which contain multiple valid solution paths.

We observe that in some cases SMRC achieves lower CSRR than other methods. This is because our framework optimizes answer correctness, while step retention is only encouraged. When these objectives conflict, SMRC prioritizes a correct final answer, which may lead to lower CSRR.

\subsection{Ablation Study}

\subsubsection{Base LM vs. Fine-tuned Reward Model}

\label{sec:base_fine}

To validate the necessity of fine-tuning the reward model, we compared several instruction-tuned base models (Qwen2.5-Instruct 7B/14B/72B and Llama-3.1-Instruct 8B/70B) used directly as reward models against our fine-tuned 14B \texttt{SMRC} model.

Results demonstrate that our fine-tuned 14B model achieves superior performance. While base models struggle to capture fine-grained process-level distinctions—showing insufficient accuracy in identifying reasoning step correctness and considerable performance volatility—our task-specific fine-tuned model provides substantially more reliable assessments of reasoning step quality, with stable and balanced performance across all metrics.

\subsubsection{Impact of the Feedback Mechanism}
\label{sec:feedback}
We conduct an ablation study on the feedback mechanism of \texttt{SMRC} across iterations on two benchmark datasets (Fig.~\ref{fig:a1}). Performance improves with iterations and converges after 3–4 rounds. This suggests that moderate iterations balance computational cost and performance, while the increase in harmonic mean demonstrates the feedback mechanism’s effectiveness in improving correction quality and preserving correct reasoning.

\section{Conclusion}
This work introduces \texttt{SMRC} framework to address the misalignment between large language models and students' reasoning pathways in mathematical error correction. We formulate \emph{\textbf{student-aware mathematical reasoning correction}} as a dual-objective optimization problem that jointly optimizes answer correctness and preservation of students' cognitive processes, and solve it using a MCTS algorithm with guided exploration, multi-step verification, and dynamic pruning.

We further construct the \textbf{MSEB} dataset of real-world student solutions with annotated errors and introduce \textbf{Correct Step Retention Rate (CSRR)} to evaluate educational alignment. Experiments show that \texttt{SMRC} achieves strong and robust performance across multiple benchmarks.

\section*{Limitations}
Despite the strong performance of \texttt{SMRC}, several limitations remain.

First, the current uniform reward backtracking strategy may not fully capture the heterogeneous contributions of individual reasoning steps in complex reasoning processes. Designing more fine-grained credit assignment mechanisms is an interesting direction for future work.

Second, \texttt{SMRC} relies on MCTS-based search, which introduces additional computational overhead as the number of reasoning steps increases, even with pruning and reward guidance. Improving the efficiency of search-based correction frameworks remains an important direction for future research.

Third, while the MSEB dataset is constructed from authentic high school student solutions, its scale is still relatively limited due to the difficulty of collecting and annotating real student reasoning traces. Expanding the dataset with broader coverage of mathematical topics and student reasoning patterns is an important direction for future work.

Finally, some evaluations rely on LLM-based automatic judges, which may introduce model-specific biases. Future work could incorporate additional human evaluation to further strengthen the reliability of the evaluation.

\section*{Acknowledgments}
This study was funded by Guangxi Science and Technology Program (2025AB25069309), the Open Research Fund of Key Laboratory of Advanced Theory and Application in Statistics and Data Science (East China Normal University). 

\bibliography{custom}

\appendix

\section*{Appendix}
\label{app:prompts}

\section{The Use of Large Language Models (LLMs)}
The authors leveraged LLMs solely as a writing aid to improve textual expression. The conceptual development, research execution, and final validation were entirely conducted and overseen by the authors, who take full responsibility for the work's integrity.

\section{Ethical Considerations}
The data used in this study was collected in an educational setting with approval from the participating school. Students participated voluntarily and were informed that their responses would be used for research purposes. No personally identifiable information was collected or released. The collected data consists solely of task-oriented mathematical problem-solving responses and poses minimal risk to participants.

\section{The MSEB Dataset Example and Description}

This appendix illustrates the structure, design principles, and utility of the MSEB (Multi-Solution Error Benchmark) dataset through a concrete example, demonstrating its role in evaluating the task of \textit{Student Aware Mathematical Reasoning Correction}.

\subsection{Data Format}

Each instance in the MSEB dataset is stored as a JSON object containing the following core fields:

\begin{itemize}
    \item \texttt{"question"}: The statement of the mathematical problem.
    \item \texttt{"answer"}: The standard correct solution and complete solving process for the problem.
    \item \texttt{"student\_answer"}: A solution attempt collected from a real student, containing common errors.
    \item \texttt{"correct\_step"}: A sequence of reasoning steps from the student's attempt that are judged to be correct. This field is used to calculate the \textbf{Correct Step Retention Rate (CSRR)}.
\end{itemize}

\subsection{Data Instance}
\label{app:data_ins}

The following example showcases a complete data instance. It exemplifies a typical scenario where a student demonstrates correct initial reasoning but makes a critical error in a later transformation.

\begin{datasetbox}[MSEB Dataset Instance]
\textbf{Question:} Given real numbers $a$, $b$, $c$ such that $a + b + c = 0$ and $a^{2}+b^{2}+c^{2}=1$, the maximum value of $a$ is \_\_\_\_\_\_\_.

\medskip

\textbf{Standard Answer:} 
\begin{enumerate}
    \item Substituting $c=-(a + b)$ into $a^{2}+b^{2}+c^{2}=1$ yields $2a^{2}+2b^{2}+2ab = 1$. 
    \item That is, $2a^{2}-1=-2b(a + b) \leq 2(\frac{-b + a + b}{2})^{2}=\frac{a^{2}}{2}$. 
    \item Therefore, $a^{2} \leq \frac{2}{3}$, i.e., $-\frac{\sqrt{6}}{3}\leq a\leq \frac{\sqrt{6}}{3}$. 
    \item Hence, the maximum value of $a$ is $\frac{\sqrt{6}}{3}$.

\end{enumerate}

\medskip

\textbf{Student Answer:}
\begin{enumerate}
    \item Express $c$ as $-(a + b)$ and substitute into $a^2+b^2+c^2=1$, obtaining $a^2+b^2+(a+b)^2=1$.
    \item Expand and simplify: $a^2 + b^2 + a^2 + 2ab + b^2 = 1$; $2a^2 + 2b^2 + 2ab = 1$; $a^2 + b^2 + ab = \frac{1}{2}$.
    \item ... (subsequent erroneous reasoning) ...
    \item In conclusion, the maximum value of $a$ is $\boxed{\frac{\sqrt{6}}{6}}$.
\end{enumerate}

\medskip

\textbf{Correct Steps:}
\begin{enumerate}
    \item Express $c$ as $-(a + b)$ and substitute into $a^2+b^2+c^2=1$, obtaining $a^2+b^2+(a+b)^2=1$.
    \item Expand and simplify the equation: $a^2 + b^2 + a^2 + 2ab + b^2 = 1$; $2a^2 + 2b^2 + 2ab = 1$; $a^2 + b^2 + ab = \frac{1}{2}$.
    \item $a^2 + b^2 + ab = (a + \frac{b}{2})^2 + \frac{3}{4}b^2 = \frac{1}{2}$.
\end{enumerate}
\end{datasetbox}

\section{Dataset Coverage Analysis of MSEB}
\label{sec:mseb_analysis}

To better understand the coverage of the MSEB dataset, we analyze the distributions of question formats, knowledge areas, and error types. These statistics help illustrate the diversity of mathematical problems and reasoning errors represented in the dataset.

\subsection{Question Format Distribution}

\begin{table}[h]
\centering
\small
\begin{tabular}{lcc}
\toprule
\textbf{Question Format} & \textbf{Count} & \textbf{Percentage} \\
\midrule
Solution-based Questions & 73 & 46.20\% \\
Proof Questions & 41 & 25.95\% \\
Fill-in-the-Blank Questions & 35 & 22.15\% \\
Multiple-Choice Questions & 9 & 5.70\% \\
\bottomrule
\end{tabular}
\caption{Distribution of question formats in MSEB.}
\label{tab:mseb_format}
\end{table}

The dataset covers multiple assessment formats commonly used in high-school mathematics exams, with solution-based and proof questions forming the majority of instances.

\subsection{Knowledge Area Distribution}

\begin{table}[h]
\centering
\small
\begin{tabular}{lcc}
\toprule
\textbf{Knowledge Area} & \textbf{Count} & \textbf{Percentage} \\
\midrule
Inequalities & 38 & 24.05\% \\
Analytic Geometry & 26 & 16.46\% \\
Trigonometric Functions & 16 & 10.13\% \\
Sequences and Series & 13 & 8.23\% \\
Probability and Statistics & 8 & 5.06\% \\
Plane Geometry & 7 & 4.43\% \\
Sequences and Inequalities & 7 & 4.43\% \\
Functions and Derivatives & 14 & 8.86\% \\
Other Algebra and Function Problems & 29 & 18.35\% \\
\bottomrule
\end{tabular}
\caption{Distribution of knowledge areas in MSEB.}
\label{tab:mseb_knowledge}
\end{table}

MSEB spans a diverse set of mathematical topics commonly taught in high-school curricula, including algebra, geometry, trigonometry, and probability.

\subsection{Error Type Distribution}

\begin{table}[h]
\centering
\small
\begin{tabular}{lcc}
\toprule
\textbf{Error Type} & \textbf{Count} & \textbf{Percentage} \\
\midrule
Computational Errors & 76 & 48.10\% \\
Logical Reasoning Errors & 67 & 42.41\% \\
Formula Misuse & 5 & 3.16\% \\
Improper Bounding Method & 7 & 4.43\% \\
Other Error Types & 3 & 1.90\% \\
\bottomrule
\end{tabular}
\caption{Distribution of error types in MSEB.}
\label{tab:mseb_error}
\end{table}

The error distribution shows that computational errors and logical reasoning mistakes account for the majority of cases. This pattern reflects common error types observed in real student work rather than synthetic error distributions generated by language models.

\paragraph{Discussion}

Overall, MSEB covers a wide range of question formats, mathematical topics, and reasoning error types. Although the dataset is relatively small, it captures realistic student reasoning patterns.

In addition, \texttt{SMRC} demonstrates consistent improvements on larger benchmarks such as \textsc{ProcessBench} and \textsc{MR-GSM8K}, suggesting that the performance gains are not tied to a particular dataset distribution.

\section{\texttt{SMRC} framework node generates prompt}
\begin{datasetbox}[\texttt{SMRC} framework node generates prompt]
You are a high-precision mathematical problem-solving and error-correction engine. Your task is to receive a [Question] and a [Student's Solution Process], and generate a "minimally corrected" correct solution.

\textbf{[Core Principles]}

1.  \textbf{Identify the Method}: First, analyze and understand the problem-solving strategy chosen by the student (e.g., substitution method or elimination by addition/subtraction for systems of equations? Formula method or completing the square for quadratic equations?).

2.  \textbf{Preserve Correct Parts}: Retain, exactly as written,\textbf{all correct steps} in the student's solution process that occur\textbf{before the first error}.

3. \textbf{Correct Within the Chosen Method}: Starting from the first error, you\textbf{must continue and complete the solution using the strategy already chosen by the student}.\textbf{It is forbidden} to switch to a different solution method.

4. \textbf{Clean Output}: Directly output the complete, correct solution process without any explanations, titles, or extraneous text.

Please strictly imitate the following examples when handling the final [Formal Task].

\textbf{Example 1: Algebraic Operation Error}

\textbf{[Question]\&[Student's Solution Process]}

Solve the equation: 5(x + 1) - 2 = 23

Solution:

5(x + 1) - 2 = 23

5x + 1 - 2 = 23

5x - 1 = 23

5x = 24

x = 4.8

\textbf{[Your Output]}

Solution:

5(x + 1) - 2 = 23

5x + 5 - 2 = 23

5x + 3 = 23

5x = 20

x = 4

\textbf{Example 2: Geometric Concept Confusion}

\textbf{[Question]\&[Student's Solution Process]}

What is the area of a square inscribed in a circle with a radius of 5cm?

Solution:

The radius of the circle r = 5cm.

The diameter of the circle d = 10cm.

The side length of the square a = d = 10cm.

The area of the square S = a² = 10² = 100 cm².

\textbf{[Your Output]}

Solution:

The radius of the circle r = 5cm.

The diameter of the circle d = 10cm.

The diagonal length of this inscribed square is equal to the diameter of the circle, which is 10cm.

Let the side length of the square be a, then a² + a² = 10².

2a² = 100

a² = 50

The area of the square S = a² = 50 cm².

\textbf{Example 3: Must Adhere to the Student's Chosen Strategy}

\textbf{[Question]\&[Student's Solution Process]}

Solve the system of equations:

(1) x + y = 3

(2) 2x - y = 6

*(The student chose the "substitution method")*
Solution:

From (1), we get x = 3 + y

Substituting x into (2) gives:

2(3 + y) - y = 6

6 + 2y - y = 6

y = 0

Substituting y=0 into (1) gives x = 3.

So the solution is x=3, y=0.

\textbf{[Your Output]}

*(The model must also use the "substitution method" for correction, not switch to the simpler "elimination by addition/subtraction method")*

Solution:

From (1), we get x = 3 - y

Substituting x into (2) gives:

2(3 - y) - y = 6

6 - 2y - y = 6

6 - 3y = 6

-3y = 0

y = 0

Substituting y=0 into x = 3 - y gives x = 3.

So the solution to the system is x=3, y=0.

\end{datasetbox}
\section{Other prompts}
\subsection{Multi-turn Feedback Prompt for the \texttt{SMRC} Framework}

\begin{datasetbox}[Multi-turn Feedback Prompt for the \texttt{SMRC} Framework]
It seems there might be some issues with your answer. Please review it and provide a new response.
\end{datasetbox}

\subsection{Student Solution Decomposition Prompt}

\begin{datasetbox}[Student Solution Decomposition Prompt]
Please help me break down the steps in the student's answer and provide them in the following format:

Steps 1:... 

Steps 2:... 

Steps 3:...

The student's answer is as follows:

\{student\_answer\}
\end{datasetbox}

\section{The case of init for Monte Carlo Tree in \texttt{SMRC}}
\label{appedix:tree_case}
The figure \ref{fig:init_tree} illustrates the answer combination tree for question Q with value scores (v) computed by the Reward Model. Red dashed edges denote pruned paths where values decrease relative to parent nodes ($a_1 \rightarrow a_3$ and $Q \rightarrow a_3$). The path $a_1 \rightarrow a_2 \rightarrow a_3$ achieves the highest score (v:0.6), representing the optimal sequence.

\begin{figure}[h]
\centering
\resizebox{0.99\columnwidth}{!}{
\begin{forest}
  for tree={
    draw,
    rounded corners,
    align=center,
    minimum width=1.5cm,
    minimum height=0.6cm,
    s sep=6mm,
    l sep=10mm,
    edge={->, >=stealth, thick}
  }
  [Q(v:0)
    [a1(v:0.2)
        [a2(v:0.4)
            [a3(v:0.6)]
        ]
        [a3(v:0.1), edge={dashed, red}]
    ]
    [a2(v:0.1)
        [a3(v:0.3)]
    ]
    [a3(v:-0.1), edge={dashed, red}]
  ]
\end{forest}}
\caption{Initialization tree for Question Q with Student Answers A = \{$a_1, a_2, a_3$\}}
\label{fig:init_tree}
\end{figure}

\section{Example of Reward Data Construction}
\label{app:reward}

As illustrated in Figure \ref{fig:binary_tree}, we employ a breadth-first search (BFS) strategy to expand multiple reasoning paths for problem Q, constructing a complete search tree structure. For value assessment, leaf nodes corresponding to reasoning paths that yield correct answers are assigned a value of +1, while leaf nodes associated with reasoning paths that produce incorrect answers are assigned a value of -1.

\begin{figure}[H]
\centering
\resizebox{0.99\columnwidth}{!}{%
\begin{forest}
  for tree={
    draw,
    rounded corners,
    align=center,
    minimum width=1.5cm,
    minimum height=0.6cm,
    s sep=6mm,
    l sep=10mm,
    edge={->, >=stealth, thick}
  }
  [Q(v:0)
    [a1
      [a3
        [a7(v:1), fill=green!30]
      ]
      [a4
        [a8(v:-1), fill=red!30]
      ]
    ]
    [a2
      [a5
        [a9(v:1), fill=green!30]
        [a10(v:1), fill=green!30]
      ]
      [a6
        [a11(v:-1), fill=red!30]
      ]
    ]
  ]
\end{forest}
}
\caption{Initial reasoning tree structure with leaf node value assignments. Green nodes indicate correct answers (+1), red nodes indicate incorrect answers (-1).}

\label{fig:binary_tree}
\end{figure}

\begin{figure*}[ht]
    \centering
    \includegraphics[width=0.98\linewidth]{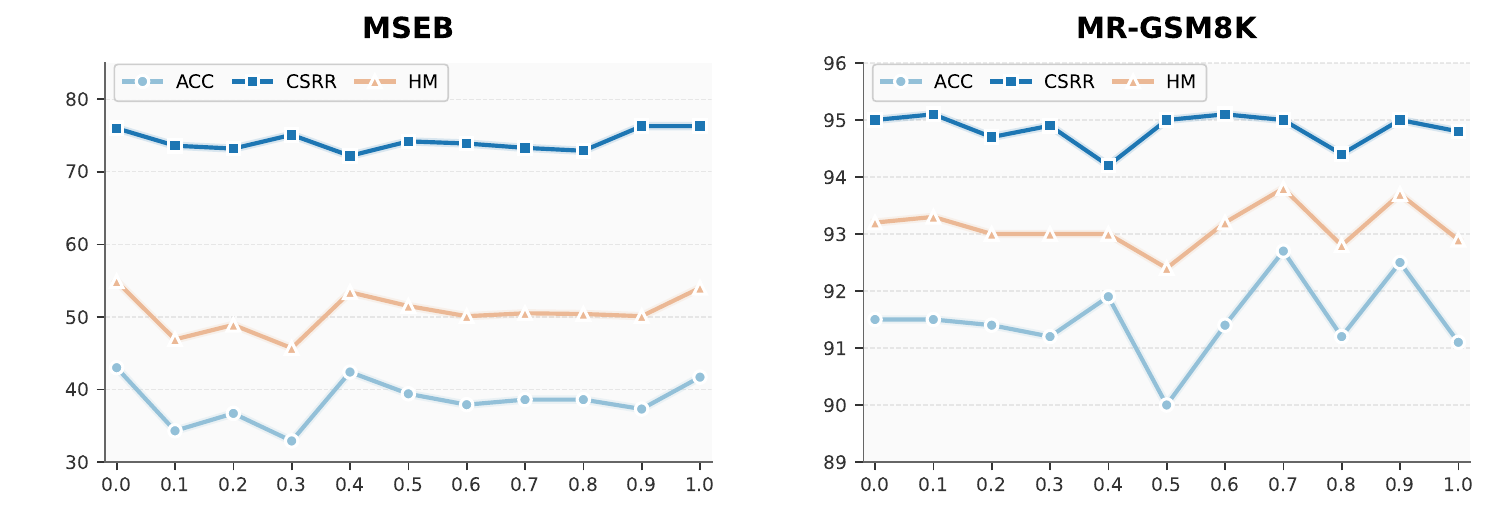}
    \caption{Effect of MCTS exploration parameter $c$ on error correction performance.}
    \label{fig:a2}
\end{figure*}

The computational workflow for step-level rewards in the reasoning tree is as follows:

\begin{enumerate}
    \item Identify the longest correct reasoning sequence (using node $a_7$ as an example), backtracking from the leaf node to the root node until the first value-annotated node is found. Nodes along the path equally distribute the total reward value $R$ of the reasoning path, with each node receiving reward:
    \begin{equation}
        r_i = \frac{R}{n}
    \end{equation}
    where $n$ is the number of nodes in the path.
    
    \item Node values are computed using cumulative calculation:
    \begin{equation}
        V(n_i) = V(\text{parent}(n_i)) + r_i
    \end{equation}
    
    \item Consequently, for a path with $R = 1$ and $n = 3$:
    \begin{align}
        V(a_1) &= 0.33 \\
        V(a_3) &= 0.33 + 0.33 = 0.66 \\
        V(a_7) &= 0.66 + 0.33 = 1.0
    \end{align}
    demonstrating the incremental contribution of the reasoning pathway.
    \item Then select the next node, $a_9$ and $a_{10}$, to apply the same approach.
    \item Finally, for incorrect reasoning paths: Select a leaf node from an erroneous path (e.g., $a_8$). Backtrack upward to locate the first value-assigned node $a_1$ with $V(a_1) = 0.33$. The path nodes share the negative contribution:
    \begin{equation}
        \begin{split}
            \text{shared contribution} &= -1 - V(a_1) \\
            &= -1 - 0.33  \\
            &= -1.33
        \end{split}
    \end{equation}
    
    Each node ($a_4$ and $a_8$) receives $r = \frac{-1.33}{2} = -0.665$. Therefore:
    \begin{equation}
        \begin{split}
            V(a_4) &= V(a_1) + r = 0.33 + (-0.665)  \\
        &= -0.335 \\
        \end{split}
    \end{equation}
    \begin{equation}
        \begin{split}
            V(a_8) &= V(a_4) + r = -0.335 + (-0.665) \\
            &= -1.0
        \end{split}
    \end{equation}
\end{enumerate}

The final constructed result is illustrated in Figure~\ref{fig:binary_tree_2}.

\begin{figure}[H]
\centering
\resizebox{0.9\columnwidth}{!}{%
\begin{forest}
  for tree={
    draw,
    rounded corners,
    align=center,
    minimum width=1.5cm,
    minimum height=0.6cm,
    s sep=6mm,
    l sep=10mm,
    edge={->, >=stealth, thick}
  }
  [Q(v:0)
    [a1(v:0.33)
      [a3(v:0.66)
        [a7(v:1), fill=green!30]
      ]
      [a4(v:-0.335)
        [a8(v:-1), fill=red!30]
      ]
    ]
    [a2(v:0.33)
      [a5(v:0.66)
        [a9(v:1), fill=green!30]
        [a10(v:1), fill=green!30]
      ]
      [a6(v:-0.335)
        [a11(v:-1), fill=red!30]
      ]
    ]
  ]
\end{forest}
}
\caption{Reasoning tree after step-level reward propagation. Node values reflect cumulative rewards computed through backtracking from leaf nodes to the root.}
\label{fig:binary_tree_2}
\end{figure}

\begin{table*}[ht]
\centering
\resizebox{0.99\textwidth}{!}{
\begin{tabular}{lcccc}
\toprule
\textbf{Method} & \textbf{Harmonic Mean} & \textbf{Avg. Total Token} & \textbf{Avg. Time / Case} & \textbf{Time Overhead} \\
\midrule
DEC & 38.3 & 1.2k & 44.88 s & 1.00$\times$ \\
TPEC & 40.8 & 1.4k & 52.53 s & 1.17$\times$ \\
Self-Check & 41.9 & 68.6k & 959.28 s & 21.37$\times$ \\
Self-Refine & 48.1 & 7.4k & 84.39 s & 1.88$\times$ \\
\textbf{SMRC} & \textbf{51.8} & 51.2k & 118.37 s & 2.64$\times$ \\
\bottomrule
\end{tabular}
}
\caption{Performance-cost trade-off on the MSEB dataset. \texttt{SMRC} achieves the best harmonic mean while incurring moderate computational overhead compared with lightweight baselines.}
\label{tab:efficiency}
\end{table*}

\section{Performance Comparison of Search Algorithms in the \texttt{SMRC} Framework: A Case Study of BFS, DFS, and MCTS}
Table~\ref{tab:search} compares the performance of three search algorithms. Although BFS and DFS achieve higher CSRR (BFS: 98.6\%/97.6\%; DFS: 96.9\%/92.0\%), they lack effective node quality assessment mechanisms and tend to retain all explored steps, resulting in lower accuracy (BFS: 15.0\%/8.86\%; DFS: 83.2\%/27.2\%).

MCTS, guided by the Reward Model for pruning, achieves slightly lower CSRR (94.5\%/73.7\%) but significantly higher accuracy (91.4\% and 40.1\%), yielding the best harmonic mean scores (92.9/51.8). This demonstrates that value-driven pruning effectively balances step retention and answer accuracy.

\begin{table}[H]
  \centering
  \resizebox{0.99\columnwidth}{!}{
  \begin{tabular}{lc|cc>{\columncolor{blue!20}}ccc>{\columncolor{blue!20}}c}
    \toprule
    & & \multicolumn{3}{c}{MR-GSM8K} & \multicolumn{3}{c}{MSEB} \\
    \midrule
    Method && ACC & CSRR & HM & ACC & CSRR & HM  \\
    \midrule
    BFS & & 15.0 & \textbf{98.6} & 26.1 & 8.86 & \textbf{97.6} & 16.2\\
    DFS & &  \underline{83.2} & \underline{96.9} & \underline{89.5} & \underline{27.2} & \underline{92.0} & \underline{42.0}\\
    \midrule
    MCTS & & \textbf{91.4} & 94.5 & \textbf{92.9} & \textbf{40.1 } & 73.7  & \textbf{51.8} \\
    \bottomrule
  \end{tabular}}
  \caption{Performance comparison of search algorithms in the \texttt{SMRC} framework on MR-GSM8K and MSEB datasets.}
  \label{tab:search}
\end{table}

\section{Efficiency Analysis}
\label{sec:efficiency}

Table~\ref{tab:efficiency} shows the performance–cost trade-off on the MSEB dataset.
SMRC achieves the best harmonic mean with about 2–3× higher runtime than lightweight baselines like DEC. Although it uses more tokens than DEC and TPEC, it is still far more efficient than Self-Check, which relies on large-scale re-sampling. Moreover, MCTS runs with a fixed rollout budget, keeping inference cost explicitly bounded.

\section{Sensitivity Analysis of MCTS Exploration Parameter}
\label{sec:c}
As shown in Fig.~\ref{fig:a2}, our framework demonstrates \textit{acceptable but limited} robustness to the exploration parameter $c$ ($c \in [0.0, 1.0]$). On MSEB, metric variation remain manageabl: ACC varies by $\sim$7\%, CSRR by $\sim$6\%, and HM by $\sim$8\%. MR-GSM8K shows better stability with $<$2\% variation across all metrics.

Despite measurable sensitivity, the framework maintains usable performance throughout the parameter range. Consistently high CSRR values confirm reliable preservation of correct reasoning steps. These results indicate that while parameter tuning yields marginal improvements, the framework remains practically applicable even with suboptimal $c$ values, reducing dependence on hyperparameter optimization.

\begin{table}[t]
  \centering
  \resizebox{\columnwidth}{!}{
  \begin{tabular}{lc|cc>{\columncolor{blue!20}}ccc>{\columncolor{blue!20}}c}
    \toprule
    & & \multicolumn{3}{c}{MR-GSM8K} & \multicolumn{3}{c}{MSEB} \\
    \midrule
    Method && ACC & CSRR & HM & ACC & CSRR & HM  \\
    \midrule
    DEC & &37.1 & 80.7 & 50.8 & 6.31 & 38.6 & 10.8\\
    TPEC & &6.30& 38.6 & 10.8 & \underline{10.8} & \underline{63.6} & \underline{18.5}\\
    Self-Check~\citep{miao2024selfcheck} & &\underline{48.5} & \underline{91.2} & \underline{63.3} & 8.01 & 60.3 & 14.4\\ 
    Self-Refine~\citep{madaan2023selfrefine}& &21.6 & 79.8 & 34.1&6.96 & \textbf{77.1} & 12.7\\
    \midrule
    \texttt{SMRC}(ours) & &\textbf{67.8} & \textbf{92.8} & \textbf{78.4} & \textbf{11.6} & 53.2 & \textbf{19.1}\\
    \bottomrule
  \end{tabular}
  }
  \caption{Comparison of \texttt{SMRC} against baseline methods on MR-GSM8K and MSEB using Llama-3.1-8B-Instruct as the generator. Best results are in \textbf{bold} and second-best are \underline{underlined}.}
  \label{tab:llama}
\end{table}

\section{Generality of \texttt{SMRC} Across Base Models}
\label{sec:gener}
As shown in Table \ref{tab:llama}, a key strength of our \texttt{SMRC} framework lies in its model-agnostic nature and consistent effectiveness across different base models. To verify that the performance improvements are attributable to our algorithmic design rather than model-specific properties, we evaluated our approach on Llama-3.1-8B, a model that possesses significantly fewer parameters than the Qwen2.5-72B model used in our main experiments. Despite this substantial reduction in model capacity, \texttt{SMRC} maintains superior performance over all baseline methods on both benchmarks. This result underscores the efficacy of its structured Monte Carlo Tree Search and external reward guidance, enabling it to achieve the highest harmonic mean scores.

\section{Implementation Details}
\label{app:implement}
In our experimental setup for the \texttt{SMRC} framework, we employ the hyperparameter configuration: the exploration parameter \(c\) for Monte Carlo Tree Search is set to 0.4, with a maximum of 30 exploration attempts and a reward threshold of 0.95 for termination. The self-correction mechanism is constrained to 4 rounds.
For model specifications, we utilize Qwen2.5-14B-Instruct fine-tuned on our reward dataset to estimate Q-values, while Qwen2.5-72B-Instruct serves as the generator for node expansion in the Monte Carlo tree.

\end{document}